\documentclass{article}
\usepackage{spconf,amsmath,graphicx,hyperref}
\usepackage{multirow}
\usepackage{amssymb}

\title{Temporal-Aware Heterogeneous Graph Reasoning with Multi-View Fusion for  Temporal Question Answering}
\name{
\begin{tabular}{c}
Wuzhenghong Wen$^{1,*}$,
Bowen Zhou$^{2,*}$,
Jinwen Huang$^{2}$,
Xianjie Wu$^{4,\dagger}$\\
Yuwei Sun$^{1}$, 
Su Pan$^{1}$,
Liang Li$^{3}$,
Jianting Liu$^{5}$
\end{tabular}%
\thanks{$^{*}$Equal contribution. $^{\dagger}$Corresponding author.}
}
\address{$^{1}$School of Internet of Things, Nanjing University of Posts and Telecommunications\\
$^{2}$Faculty of Medicine \& Health, University of New South Wales, Sydney, Australia\\
$^{3}$Dept. of Biomedical Engineering \& Biotechnology, Khalifa University, Abu Dhabi\\
$^{4}$Beijing Information Science and Technology University \quad
$^{5}$Shortest Path Technology Co., Ltd.}

\begin{document}
%
\maketitle
\begin{abstract}
Question Answering over Temporal Knowledge Graphs (TKGQA) has attracted growing interest for handling time-sensitive queries. However, existing methods still struggle with: 1) weak incorporation of temporal constraints in question representation, causing biased reasoning; 2) limited ability to perform explicit multi-hop reasoning; and 3) suboptimal fusion of language and graph representations.
We propose a novel framework with temporal-aware question encoding, multi-hop graph reasoning, and multi-view heterogeneous information fusion. Specifically, our approach introduces: 1) a constraint-aware question representation that combines semantic cues from language models with temporal entity dynamics; 2) a temporal-aware graph neural network for explicit multi-hop reasoning via time-aware message passing; and 3) a multi-view attention mechanism for more effective fusion of question context and temporal graph knowledge.
Experiments on multiple TKGQA benchmarks demonstrate consistent improvements over multiple baselines.
\end{abstract}
\section{Introduction}
\label{sec:intro}

Recently, Temporal Knowledge Graph Question Answering (TKGQA) has emerged as a promising research direction, exhibiting immense potential in real-world applications. Traditional KGQA and TKGQA differ significantly in several aspects: (1) TKGQA handles more complex queries due to the inclusion of temporal scopes. While traditional KGs represent facts as (subject, predicate, object) triples, temporal KGs extend this structure to (subject, predicate, object, temporal information). (2) Answers in TKGQA can be of diverse types. For instance, a fact such as (LeBron James, playFor, Miami Heat, [2010, 2014]) enables TKGQA systems to return either entity-based answers (e.g., “Miami Heat”) or temporal intervals (e.g., “2010 to 2014”). This diversity necessitates additional temporal reasoning capabilities.

Existing approaches address TKGQA by decomposing questions into temporal and non-temporal sub-questions \cite{jia2018tequila}, or by integrating pre-trained models (PLMs) with temporal KGs for direct answer generation \cite{saxena2021question}. While these methods perform adequately on simple questions involving single entities or explicit time points, they often fail on queries with complex constraints and multi-hop structures, which involves both constrained bias and multi-hop reasoning, commercial systems like Google Search fail to capture entity changes due to temporal constraints, and ChatGPT suffers from hallucination issues \cite{sun2023contrastive,liu2023local,song2022improving}. We argue that current solutions do not adequately address the core challenges of TKGQA \cite{Tan2023TowardsBA,wu2025tablebench}, particularly the following aspects:
\textbf{Q1: How to perceive entity changes caused by constraints during question representation?}
Existing methods often encode questions using PLMs without leveraging associated temporal knowledge from the knowledge graph. This overlooks rich temporal information related to question entities, causing models to overfit to explicit entity mentions and ignore implicit entity transitions due to constraints. Consider the question \textit{“Who was Lasha Talakhadze’s previous Olympic coach?”}. Static encoding ignores the temporal constraint “previous,” while a temporal-aware model could use the KG to detect changes in coaching relationships over time.
\textbf{Q2: How to explicitly answer questions that require multi-hop reasoning over temporal facts?}
Most existing methods do not explicitly model the graph structure between entities, limiting their ability to perform multi-hop reasoning. And The answering the third question requires first retrieving the hosting time of the 18th World Cup and then determining the president during that period. Explicit multi-hop reasoning over temporal knowledge graphs is essential for such cases but remains under-explored \cite{Dong2023HierarchyAwareMQ}.
\textbf{Q3: How to effectively integrate heterogeneous representations from language and knowledge graphs?}
Representations from PLMs, KG embeddings, and graph neural networks often lie in different vector spaces. Prior methods rely on simple concatenation or weighted aggregation, which may lead to suboptimal fusion. 

To tackle these issues, we proposed Temporal-Aware Heterogeneous Graph Reasoning with Multi-View Fusion method which designed to mirror human-like progressive reasoning and consists of three stages:
(i) a \textit{constraint-aware question understanding} module that enhances semantic question representation with temporal KG context;
(ii) a \textit{temporal-aware graph neural network} that supports multi-hop reasoning with time-aware message propagation; and
(iii) a \textit{multi-view attention mechanism} that aligns and fuses heterogeneous representations from both language and graph domains.

The main contributions of this work are summarized as follows: we introduce a temporal-aware graph neural network that can capture both structural and temporal dependencies in multi-hop reasoning paths; we design a multi-view attention module that enables the adaptive fusion of heterogeneous representations from language models and knowledge graphs; and we conduct extensive experiments on multiple Temporal Knowledge Graph Question Answering (TKGQA) benchmark datasets, with results demonstrating that the proposed model achieves consistent improvements over the baselines.

\section{RELATED WORK}
\label{sec:format}
Generally, KG embedding algorithms \cite{garcia2018learning,gao2025decorl} are used to initialize entity and relation embeddings in KGQA \cite{xue2024question,liu2024resolving,xue2023dual}. For TKGQA, temporal KG embeddings such as TComplEx \cite{li2024local} are typically adopted to also obtain timestamp embeddings. Recent works have proposed various TKGQA models \cite{wu2025progressive,li2024comateformer}, among which three are representative: CronKGQA \cite{saxena2021question,wu2025unleashing}, TMA \cite{liu2023time}, and TSQA \cite{shang2022improving,wang2025not,wu2025breaking}. CronKGQA leverages temporal KG embeddings and pre-trained LMs for answer prediction. TMA extracts entity-related information from questions and fuses heterogeneous signals via a multi-view mechanism.

\noindent \textbf{Temporal KGQA}:
\noindent \textbf{Graph Neural Networks}: GNNs have attracted tremendous attention due to their capability to model structured data and have been developed for various applications in practice \cite{liang2019adaptive,wang2022dabert,fei2022cqg,liang2019asynchronous}. Graph convolutional network (GCN) is a pioneering work that designs a local spectral graph convolutional layer for learning node embeddings. GraphSAGE  generates nodes' embeddings by learning an aggregator function that samples and aggregates feature from nodes' local neighborhoods. Graph Attention Network (GAT) \cite{liu2025structural,liu2025stole} assigns different weights to different neighbors of a node to learn its representations by introducing self-attention mechanisms. Recently, several models \cite{zheng2022robust,gui2018transferring,ma2022searching,dai2025hope} have been designed to shift the power of GNNs to general QA tasks. 
Nevertheless, these models merely use vanilla GNNs that adopt a one-hop neighbor aggregation mechanism, which may limit their expressiveness. 
Meanwhile, these models cannot be directly applied to our focused scenarios, \textit{i.e.}, TKGQA. 

\section{Problem Definition}
\label{sec:pagestyle}
\noindent\textbf{TKGQA.} 
 When dealing with a new question $q$ from the user, the model needs to identify correct temporal $t$ or entity nodes $n$ in order to provide an accurate answer from a temporal KG $\mathcal{G}=(\mathcal{E}, \mathcal{P}, \mathcal{T}, \Upsilon)$ comprised of tuple such as $(s, p, o, t)$.
A TKG represents such a fact as a quadruple of the form $\{s,p,o,t\}$, where $t_{st}$ is the start time and $t_{et}$ is the end time of validity of the fact, 
example.
\noindent\textbf{Temporal KG embedding}
can learn representation of $e_s$, $e_p$, $e_o$, $e_t$ $\in$ $\mathbb{R}^{d}$ for each $s, o \in \mathcal{E}, p \in \mathcal{P}, t \in \mathcal{T}$. 
Intuitively, we can establish a scoring function $\phi(\cdot)$, which employs semantic similarity to item embeddings.
For a accurate fact $\upsilon=(s, p, o, t) \in \Upsilon$, it will be scored higher than an erroneous fact $\upsilon^{\prime}=(s^\prime, p^\prime, o^\prime, t^\prime) \notin \Upsilon$ via the function $\phi(\cdot)$. 

\section{Method}
\label{sec:method}
\subsection{Temporal-Aware KG Embedding}

We employ a temporal knowledge graph embedding method based on TComplEx to obtain representations for entities, relations, and timestamps. The scoring function is defined as:
\begin{footnotesize}
\begin{equation}
\begin{aligned}
\phi(e_s, e_r, \bar{e}_o, e_t) &= \mathbf{Re}(\langle e_s, e_r \odot e_t, \bar{e}_o\rangle) \\
&= \mathbf{Re}\left(\sum\nolimits_{d=1}^{2D}e_s[d]e_r[d]e_t[d]\bar{e}_o[d]\right)
\end{aligned}  
\end{equation}
\end{footnotesize}
where $d$ denotes each dimension position, $D$ represents the embedding dimension, and the total dimension is $2D$ to account for both real and imaginary parts in complex space.
To explicitly incorporate temporal order information, we enhance timestamp embeddings with learnable positional encodings inspired by transformer architectures. For the $k$-th timestamp, its positional embedding $\mathbf{t}_{k}$ is defined as:
\begin{footnotesize}
\begin{equation}
\label{position}
\mathbf{t}_k(c) = 
\begin{cases}
\sin(k/10000^{2i/2d}), & \text{if } c=2i \\
\cos(k/10000^{2i/2d}), & \text{if } c=2i+1
\end{cases}
\end{equation}
\end{footnotesize}
We introduce an auxiliary training objective that predicts temporal ordering between timestamp pairs $(m, n)$:
\begin{footnotesize}
\begin{equation}
\begin{aligned}
\rho(m, n) &= \sigma(\mathbf{W}_{ts}^\top((e_m+\mathbf{t}_m) - (e_n+\mathbf{t}_n))) \\
\mathcal{L}_{ts}(m,n) &= -\alpha(m,n)\log(\rho(m,n)) \\
&-(1-\alpha(m,n))\log(1-\rho(m,n))
\end{aligned}
\end{equation}
\end{footnotesize}
where $\mathbf{W}_{ts}$ is a learnable matrix, $e_\ast$ and $\mathbf{t}_\ast$ denote timestamp and position embeddings respectively, and $\alpha(m, n)=1$ if $m < n$ and 0 otherwise. The final training objective combines both KG reconstruction and temporal ordering losses: $\mathcal{L}_{fin}=\mathcal{L}_{tc}+\lambda\mathcal{L}_{ts}$, where $\lambda$ is a balancing coefficient.

\subsection{Constraint-Aware Question Representation}

To capture entity transitions under temporal constraints, we first retrieve related SPO facts from the KG and encode both the question and facts with a pre-trained language model (PLM):  
\begin{footnotesize}
\begin{equation}
\mathbf{\mathcal{Q}} = \mathrm{PLM}(\textit{Question}), \quad 
\mathbf{s_i} = \mathrm{PLM}(\mathrm{SPO_i})
\end{equation}
\end{footnotesize}
Let $Q=[q_1,\dots,q_n]$ be token embeddings of the question and $S=[s_1,\dots,s_m]$ the [CLS] embeddings of $m$ SPOs. We apply cross-attention:
\begin{footnotesize}
\begin{equation}
\bar{Q} = \mathrm{softmax}\!\left(\tfrac{\mathcal{S}Q^T}{\sqrt{d_k}}\right)\mathcal{Q}, \quad
\bar{S} = \mathrm{softmax}\!\left(\tfrac{\mathcal{Q}S^T}{\sqrt{d_k}}\right)\mathcal{S}
\end{equation}
\end{footnotesize}
Aligned features are fused by a gating mechanism:
\begin{footnotesize}
\begin{equation}
g_i = \sigma(W_{g_i}(\bar{q_i} \oplus \bar{s_i})), \quad
v_i = g_i \bar{q_i} + (1-g_i)\bar{s_i}
\end{equation}
\end{footnotesize}
yielding the final constraint-aware representation 
$Q_{new}=[v_1,\dots,v_n]$.

\subsection{Multi-Hop Temporal Graph Reasoning}

We construct a temporal subgraph $G=(\mathcal{V}, E)$ centered on question entities. To enable explicit multi-hop reasoning, we propose a temporal-aware graph neural network with path-aware attention.
The message passing mechanism with temporal attention is formulated as:
\begin{footnotesize}
\begin{equation}
\label{attention}
{A}_{irj}^\ell =
\begin{cases}
\beta^\ell \mathrm{ReLU}(\mathbf{W}_{ad}^\ell(h_i^\ell||h_j^\ell||h_r||h_t)), & (v_i, r, v_j, t) \in G \\
-\infty, & \text{otherwise}
\end{cases}
\end{equation}

\begin{equation}
\mathbf{A}^\ell = \mathrm{softmax}(\mathbf{A}^\ell) 
\end{equation}
\end{footnotesize}
To capture multi-hop dependencies within single propagation layers, we employ a diffusion operator that aggregates information across various path lengths:
\begin{footnotesize}
\begin{equation}
\label{matrix}
\mathbf{D} = \sum\nolimits_{\tau=0}^{\aleph}\xi_\tau\mathbf{A}^{\tau}
\end{equation}
\end{footnotesize}
where $\xi_\tau$ are learnable coefficients weighting different hop distances. Node representations are updated as:
\begin{footnotesize}
\begin{equation}
\mathbf{H}^{\ell+1} = \mathbf{D}\mathbf{H^\ell}
\end{equation}
\end{footnotesize}
The final graph representation is obtained through attentive pooling over question-relevant nodes:
\begin{footnotesize}
\begin{equation}
\label{ba}
\mathbf{S}_{graph} = \frac{1}{|V_q|}\sum\nolimits_{i\in V_q}\alpha_i h^L_i
\end{equation}
\end{footnotesize}
where attention weights $\alpha_i$ are computed based on node relevance to the question.

\subsection{Multi-View Adaptive Fusion}

We propose a multi-view attention mechanism to effectively integrate heterogeneous representations from question encoding and graph reasoning.

\noindent
\textbf{View 1: Semantic-Symbolic Alignment}. We first align question and graph representations through cross-modal attention:
\begin{footnotesize}
\begin{equation}
\begin{aligned}
\mathbf{H}^{q2g} &= \mathrm{Attention}(\mathbf{Q}_{new}, \mathbf{S}_{graph}, \mathbf{S}_{graph}) \\
\mathbf{H}^{g2q} &= \mathrm{Attention}(\mathbf{S}_{graph}, \mathbf{Q}_{new}, \mathbf{Q}_{new})
\end{aligned}
\end{equation}
\end{footnotesize}
\noindent\textbf{View 2: Temporal-Aware Fusion}. We incorporate temporal information into the fusion process:
\begin{footnotesize}
\begin{equation}
\mathbf{H}^{temp} = \mathrm{ReLU}(\mathbf{W}_t[\mathbf{H}^{q2g}; \mathbf{H}^{g2q}; \mathbf{T}])
\end{equation}
\end{footnotesize}
where $\mathbf{T}$ represents temporal features derived from timestamp embeddings.

\noindent\textbf{View 3: Context-Gated Fusion}. We employ a gating mechanism to adaptively combine information from different views:
\begin{footnotesize}
\begin{equation}
\begin{aligned}
g &= \sigma(\mathbf{W}_g[\mathbf{H}^{q2g}; \mathbf{H}^{g2q}; \mathbf{H}^{temp}]) \\
\mathbf{H}^{fusion} &= g \cdot \mathbf{H}^{q2g} + (1-g) \cdot \mathbf{H}^{g2q} + \mathbf{H}^{temp}
\end{aligned}
\end{equation}
\end{footnotesize}
The final representation is used for answer prediction through a linear layer and softmax activation:
\begin{footnotesize}
\begin{equation}
P(ans|Q, G) = \mathrm{softmax}(\mathbf{W}_o\mathbf{H}^{fusion} + b_o)
\end{equation}
\end{footnotesize}
The model is trained end-to-end with cross-entropy loss combined with the temporal KG embedding objectives.

\begin{table*}[h]
	\centering
 \renewcommand\arraystretch{0.80}
	\caption{Performance of baselines and our methods on the CronQuestions dataset.}
	\label{results1} 
	\scalebox{0.85}{
        \setlength{\tabcolsep}{3.7mm}{
	\begin{tabular}{l|c|cc|cc|c|cc|cc}
	    \hline
		\multirow{3}*{\textbf{Model}}& \multicolumn{5}{c|}{\textbf{Hits@1}} & \multicolumn{5}{c}{\textbf{Hits@10}} \\
		\cline{2-11}
        ~& \multirow{2}*{\textbf{Overall}} & \multicolumn{2}{c|}{\textbf{Question Type}} & \multicolumn{2}{c|}{\textbf{Answer Type}} & \multirow{2}*{\textbf{Overall}}& \multicolumn{2}{c|}{\textbf{Question Type}} & \multicolumn{2}{c}{\textbf{Answer Type}} \\
        \cline{3-6}\cline{8-11}
        ~&~& \textbf{Complex} & \textbf{Simple} & \textbf{Entity}&\textbf{Time}&~&\textbf{Complex} & \textbf{Simple} & \textbf{Entity}&\textbf{Time}\\
        \hline
        BERT & 0.071 & 0.086 & 0.052 & 0.077 & 0.06 & 0.213 & 0.205 & 0.225 & 0.192 & 0.253 \\
        RoBERTa & 0.07 & 0.086 & 0.05 & 0.082 & 0.048 & 0.202 & 0.192 & 0.215 & 0.186 & 0.231 \\
        KnowBERT & 0.07 & 0.083 & 0.051 & 0.081 & 0.048 & 0.201 & 0.189 & 0.217 & 0.185 &0.23\\
        T5-3B&0.081 & 0.073 &0.091 & 0.088 & 0.067&-&-&-&-&- \\
        \hline
        EmbedKGQA & 0.288 & 0.286 & 0.29 & 0.411 & 0.057 & 0.672 & 0.632 & 0.725 & 0.85& 0.341\\
        T-EaE-add& 0.278 & 0.257 & 0.306 & 0.313& 0.213& 0.663 &0.614 & 0.729 & 0.662 & 0.665\\
        T-EaE-replace & 0.288 & 0.257 & 0.329 & 0.318 & 0.231 & 0.678 & 0.623 & 0.753 & 0.668 & 0.698\\
        CronKGQA & 0.647 & 0.392 & 0.987 &0.699 & 0.549 & 0.884&0.802&0.992 & 0.898 & 0.857\\
        TMA  & 0.784 & 0.632 & 0.987 & 0.792 & 0.743 & 0.943 & 0.904 & 0.995 & 0.947 & 0.936\\
        TSQA  & 0.831 & 0.713 & 0.987	& 0.829	& 0.836	& 0.980	& 0.968	& 0.997	& 0.981	& 0.978 \\
        TempoQR  & 0.918 & 0.864 & 0.990 & 0.926 & 0.903 & 0.978 & 0.967 & 0.993 & 0.980 & 0.974\\
        CTRN  & 0.920 & 0.869 & 0.990 & 0.921 & 0.917 & 0.980 & 0.970 & 0.993 & 0.982 & 0.976\\
        \hline
        \textbf{Ours} & \textbf{0.969}  & \textbf{0.962}  & \textbf{0.997}  & \textbf{0.972}  & \textbf{0.974}  & \textbf{0.992}  & \textbf{0.991}  & \textbf{0.996}  & \textbf{0.995}  & \textbf{0.990}\\ \hline
	\end{tabular}}}
\end{table*}

\section{Experiement}
\label{sec:typestyle}
\subsection{Datasets and Baselines}
\textbf{Dataset.} We employ two TKGQA benchmarks,
CRONQUESTIONS and Time-Questions .
\textbf{CRONQUESTIONS} has 125K entities, 1.7K timestamps, 203 relations and 328K facts. 
\textbf{TimeQuestions} boasting a collection of 16,000 meticulously labeled temporal question that have been thoughtfully categorized into four distinct classes.

\noindent \textbf{Baselines.} We have opted for three distinct baseline. (I) PLMs includes BERT, KnowBERT. (II) KG embedding approaches includes EmbedKGQA and CronKGQA \cite{saxena2021question}. (III) Temporal KG embedding-based models, including CronKGQA, TMA, TSQA , TempoQR and CTRN.

\subsection{Results and Analysis}
\label{sec:experiments}
We conduct comprehensive experiments to evaluate the effectiveness of our proposed Temporal-Aware Heterogeneous Graph Reasoning framework. Table~\ref{results1} and Table~\ref{result2} present the main results on CronQuestions and TimeQuestions datasets, respectively. Our model achieves state-of-the-art performance across all evaluation metrics, demonstrating significant improvements over existing methods. 
Specifically, on CronQuestions, our approach achieves 0.969 Hits@1 overall, outperforming the strongest baseline CTRN by 4.9 percentage points. 
\begin{table}[h]
	\centering
     \renewcommand\arraystretch{0.90}
	\caption{Hits@1 for different models on TimeQuestions.}
	\label{result2}  
	\resizebox{0.48\textwidth}{!}{
	\begin{tabular}{l|c|c c|c c}
	    \hline

        {\textbf{Model}}& {\textbf{Overall}} & {\textbf{Explicit}} & {\textbf{Implicit}} & {\textbf{Temporal}} & {\textbf{Ordinal}} \\

        \hline
        CronKGQA &0.393 &0.388 &0.380 &0.436 &0.332	\\
        TMA  &0.435 &0.442 &0.419 &0.476 &0.352 \\
        TempoQR &0.459 &0.503 &0.442 &0.458 &0.367\\
        CTRN  &0.466 &0.469 &0.446 &0.512 &0.382\\
        \textbf{Ours}  &\textbf{0.539} &\textbf{0.534} &\textbf{0.510} &\textbf{0.617} &\textbf{0.412}\\
        \hline
	\end{tabular}
	}
\end{table}
The improvements are particularly pronounced for complex questions (9.3\% improvement over CTRN) and entity-type answers (5.1\% improvement), indicating our model's enhanced capability in handling sophisticated temporal reasoning. On TimeQuestions, our method achieves 0.539 Hits@1, substantially exceeding CTRN's 0.466, with particularly notable gains on temporal questions (20.5\% improvement) and ordinal questions (7.9\% improvement). These results validate the effectiveness of our constraint-aware representation learning, multi-hop temporal reasoning, and multi-view fusion mechanisms in capturing complex temporal relationships and constraints inherent in real-world questions. The consistent performance gains across different question types and answer categories suggest that our approach provides a comprehensive solution to the challenge in TKGQA.
These results validate the effectiveness of our constraint-aware representation learning, multi-hop temporal reasoning, and multi-view fusion mechanisms in capturing complex temporal relationships and inherent constraints.

\vspace{-0.4cm}
\begin{table}[h]
	\centering
      \renewcommand\arraystretch{0.90}
	\caption{Results of component ablation experiment.}
	\label{result_abs}  
	\resizebox{0.48\textwidth}{!}{
	\begin{tabular}{l|c|c c|c c}
	    \hline
	    \multirow{3}*{\textbf{Model}}& \multicolumn{5}{c}{\textbf{Hits@1}}\\
		\cline{2-6}
        ~& \multirow{2}*{\textbf{Overall}} & \multicolumn{2}{c|}{\textbf{Question Type}} & \multicolumn{2}{c}{\textbf{Answer Type}} \\
        \cline{3-6}
        ~&~& \textbf{Complex} & \textbf{Simple} & \textbf{Entity}&\textbf{Time}\\
        \hline
        Ours        &0.969	&0.962	&0.997	&0.972	&0.974\\
        w/o time-aware &0.904 &0.868 &0.960 &0.875 &0.963\\
w/o adaptive  &0.890 &0.845 &0.945 &0.853 &0.951\\
w/o multi-hop &0.886 &0.813 &0.984 &0.880 &0.902\\
w/o constraint-aware &0.853 &0.768 &0.961 &0.826 &0.901\\
        \hline
	\end{tabular}
	}
\end{table}
\vspace{-0.1cm}

\subsection{Ablation Studies}
\label{sec:ablation}
We perform ablation studies to examine the contribution of each component in our framework (Table~\ref{result_abs}).
Removing the temporal-aware embedding results in a 6.5\% performance decrease, confirming the necessity of explicit time modeling. Eliminating adaptive fusion causes a 7.9\% drop, underscoring the importance of heterogeneous information integration. Without multi-hop reasoning, performance declines by 8.3\%, demonstrating the vital role of multi-step reasoning paths. The constraint-aware component proves most critical, with its removal leading to an 11.6\% performance degradation. These results validate the importance of each proposed component and their synergistic combination in our framework.

\section{Conclusion}
\label{sec:conclusion}
In this paper, we propose a \textbf{novel Temporal-Aware Heterogeneous Graph Reasoning} framework for temporal knowledge graph question answering. Our approach integrates constraint-aware question representation, multi-hop temporal reasoning, and adaptive information fusion to effectively address the unique challenges in TKGQA. Extensive experimental results show that the method has achieved consistent improvements compared with multiple  baselines. Ablation studies further confirm the importance of each component and their synergistic combination. Overall, the proposed method provide a practical solution for handling complex temporal constraints and multi-hop reasoning in temporal QA.


\bibliographystyle{IEEEbib}
\bibliography{strings,refs}

@inproceedings{saxena2021question,
  title={Question Answering Over Temporal Knowledge Graphs},
  author={Saxena, Apoorv and Chakrabarti, Soumen and Talukdar, Partha},
  booktitle={Proceedings of the Annual Meeting of the ACL and  IJCNLP},
  pages={6663--6676},
  year={2021}
}

@article{wang2022dabert,
  title={Dabert: Dual attention enhanced bert for semantic matching},
  author={Wang, Sirui and Liang, Di and Song, Jian and Li, Yuntao and Wu, Wei},
  journal={arXiv preprint arXiv:2210.03454},
  year={2022}
}

@inproceedings{liang2019asynchronous,
  title={Asynchronous deep interaction network for natural language inference},
  author={Liang, Di and Zhang, Fubao and Zhang, Qi and Huang, Xuan-Jing},
  booktitle={Proceedings of the 2019 Conference on Empirical Methods in Natural Language Processing and the 9th International Joint Conference on Natural Language Processing (EMNLP-IJCNLP)},
  pages={2692--2700},
  year={2019}
}

@inproceedings{liu2023time,
  title={Time-aware multiway adaptive fusion network for temporal knowledge graph question answering},
  author={Liu, Yonghao and Liang, Di and Fang, Fang and Wang, Sirui and Wu, Wei and Jiang, Rui},
  booktitle={ICASSP 2023-2023 IEEE International Conference on Acoustics, Speech and Signal Processing (ICASSP)},
  pages={1--5},
  year={2023},
  organization={IEEE}
}

@article{ma2022searching,
  title={Searching for optimal subword tokenization in cross-domain ner},
  author={Ma, Ruotian and Tan, Yiding and Zhou, Xin and Chen, Xuanting and Liang, Di and Wang, Sirui and Wu, Wei and Gui, Tao and Zhang, Qi},
  journal={arXiv preprint arXiv:2206.03352},
  year={2022}
}

@article{zheng2022robust,
  title={Robust lottery tickets for pre-trained language models},
  author={Zheng, Rui and Bao, Rong and Zhou, Yuhao and Liang, Di and Wang, Sirui and Wu, Wei and Gui, Tao and Zhang, Qi and Huang, Xuanjing},
  journal={arXiv preprint arXiv:2211.03013},
  year={2022}
}

@article{Tan2023TowardsBA,
  title={Towards Benchmarking and Improving the Temporal Reasoning Capability of Large Language Models},
  author={Qingyu Tan and Hwee Tou Ng and Lidong Bing}
}

@article{li2024comateformer,
  title={Comateformer: Combined Attention Transformer for Semantic Sentence Matching},
  author={Li, Bo and Liang, Di and Zhang, Zixin},
  journal={arXiv preprint arXiv:2412.07220},
  year={2024}
}

@article{Dong2023HierarchyAwareMQ,
  title={Hierarchy-Aware Multi-Hop Question Answering over Knowledge Graphs},
  author={Junnan Dong and Qinggang Zhang and Xiao Huang and Keyu Duan and Qiaoyu Tan and Zhimeng Jiang},
  journal={Proceedings of the ACM Web Conference 2023},
  year={2023}
}

@inproceedings{gui2018transferring,
  title={Transferring from formal newswire domain with hypernet for twitter pos tagging},
  author={Gui, Tao and Zhang, Qi and Gong, Jingjing and Peng, Minlong and Liang, Di and Ding, Keyu and Huang, Xuan-Jing},
  booktitle={EMNLP},
  pages={2540--2549},
  year={2018}
}

@inproceedings{garcia2018learning,
  title={Learning Sequence Encoders for Temporal Knowledge Graph Completion},
  author={Garcia-Duran, Alberto and Duman{\v{c}}i{\'c}, Sebastijan and Niepert, Mathias},
  booktitle={EMNLP}
}

@inproceedings{sun2023contrastive,
  title={Contrastive learning reduces hallucination in conversations},
  author={Sun, Weiwei and Shi, Zhengliang and Gao, Shen and Ren, Pengjie and de Rijke, Maarten and Ren, Zhaochun}
}

@inproceedings{xue2024question,
  title={Question Calibration and Multi-Hop Modeling for Temporal Question Answering},
  author={Xue, Chao and Liang, Di and Wang, Pengfei and Zhang, Jing},
  booktitle={Proceedings of the AAAI Conference on Artificial Intelligence},
  year={2024}
}

@inproceedings{liu2023local,
  title={Local and global: temporal question answering via information fusion},
  author={Liu, Yonghao and Liang, Di and Li, Mengyu and Giunchiglia, Fausto and Li, Ximing and Wang, Sirui and Wu, Wei and Huang, Lan and Feng, Xiaoyue and Guan, Renchu},
  booktitle={Proceedings of the Thirty-Second International Joint Conference on Artificial Intelligence},
  pages={5141--5149},
  year={2023}
}

@inproceedings{jia2018tequila,
  title={Tequila: Temporal question answering over knowledge bases},
  author={Jia, Zhen and Abujabal, Abdalghani and Saha Roy, Rishiraj and Str{\"o}tgen, Jannik and Weikum, Gerhard},
  booktitle={CIKM}
}

@article{liu2024resolving,
  title={Resolving Word Vagueness with Scenario-guided Adapter for Natural Language Inference},
  author={Liu, Yonghao and Li, Mengyu and Liang, Di and Li, Ximing and Giunchiglia, Fausto and Huang, Lan and Feng, Xiaoyue and Guan, Renchu},
  journal={arXiv preprint arXiv:2405.12434},
  year={2024}
}

@inproceedings{li2024local,
  title={Local and Global: Text Matching Via Syntax Graph Calibration},
  author={Li, Liang and Liao, Qisheng and Lai, Meiting and Liang, Di and Liang, Shangsong},
  booktitle={ICASSP 2024-2024 IEEE International Conference on Acoustics, Speech and Signal Processing (ICASSP)},
  pages={11571--11575},
  year={2024},
  organization={IEEE}
}

@inproceedings{shang2022improving,
  title={Improving Time Sensitivity for Question Answering over Temporal Knowledge Graphs},
  author={Shang, Chao and Wang, Guangtao and Qi, Peng and Huang, Jing},
  booktitle={ACL},
  year={2022}
}

@article{gao2025decorl,
  title={DeCoRL: Decoupling Reasoning Chains via Parallel Sub-Step Generation and Cascaded Reinforcement for Interpretable and Scalable RLHF},
  author={Gao, Ziyuan and Liang, Di and Wu, Xianjie and Morel, Philippe and Peng, Minlong},
  journal={arXiv preprint arXiv:2511.19097},
  year={2025}
}

@inproceedings{wu2025tablebench,
  title={Tablebench: A comprehensive and complex benchmark for table question answering},
  author={Wu, Xianjie and Yang, Jian and Chai, Linzheng and Zhang, Ge and Liu, Jiaheng and Du, Xeron and Liang, Di and Shu, Daixin and Cheng, Xianfu and Sun, Tianzhen and others},
  booktitle={AAAI},
  volume={39},
  number={24},
  pages={25497--25506},
  year={2025}
}

@inproceedings{liang2019adaptive,
  title={Adaptive multi-attention network incorporating answer information for duplicate question detection},
  author={Liang, Di and Zhang, Fubao and Zhang, Weidong and Zhang, Qi and Fu, Jinlan and Peng, Minlong and Gui, Tao and Huang, Xuanjing},
  booktitle={Proceedings of the 42nd international ACM SIGIR conference on research and development in information retrieval},
  pages={95--104},
  year={2019}
}

@inproceedings{fei2022cqg,
  title={CQG: A simple and effective controlled generation framework for multi-hop question generation},
  author={Fei, Zichu and Zhang, Qi and Gui, Tao and Liang, Di and Wang, Sirui and Wu, Wei and Huang, Xuan-Jing},
  booktitle={ACL},
  pages={6896--6906},
  year={2022}
}

@inproceedings{wu2025breaking,
  title={Breaking Size Barrier: Enhancing Reasoning for Large-Size Table Question Answering},
  author={Wu, Xianjie and Liang, Di and Yang, Jian and Cheng, Xianfu and Chai, LinZheng and Li, Tongliang and Yang, Liqun and Li, Zhoujun},
  booktitle={DASFAA},
  pages={241--256},
  year={2025},
  organization={Springer}
}

@article{song2022improving,
  title={Improving semantic matching through dependency-enhanced pre-trained model with adaptive fusion},
  author={Song, Jian and Liang, Di and Li, Rumei and Li, Yuntao and Wang, Sirui and Peng, Minlong and Wu, Wei and Yu, Yongxin},
  journal={arXiv preprint arXiv:2210.08471},
  year={2022}
}

@inproceedings{xue2023dual,
  title={Dual path modeling for semantic matching by perceiving subtle conflicts},
  author={Xue, Chao and Liang, Di and Wang, Sirui and Zhang, Jing and Wu, Wei},
  booktitle={ICASSP},
  year={2023},
  organization={IEEE}
}

@inproceedings{wu2025unleashing,
  title={Unleashing potential of evidence in knowledge-intensive dialogue generation},
  author={Wu, Xianjie and Yang, Jian and Li, Tongliang and Zhang, Shiwei and Du, Yiyang and Chai, LinZheng and Liang, Di and Li, Zhoujun},
  booktitle={ICASSP},
  year={2025},
  organization={IEEE}
}

@article{wu2025progressive,
  title={Progressive Mastery: Customized Curriculum Learning with Guided Prompting for Mathematical Reasoning},
  author={Wu, Muling and Qian, Qi and Liu, Wenhao and Wang, Xiaohua and Huang, Zisu and Liang, Di and Miao, LI and Dou, Shihan and Lv, Changze and Wang, Zhenghua and others},
  journal={arXiv preprint arXiv:2506.04065},
  year={2025}
}

@article{wang2025not,
  title={Not all parameters are created equal: Smart isolation boosts fine-tuning performance},
  author={Wang, Yao and Liang, Di and Peng, Minlong},
  journal={arXiv preprint arXiv:2508.21741},
  year={2025}
}

@inproceedings{liu2025structural,
  title={Structural Reward Model: Enhancing Interpretability, Efficiency, and Scalability in Reward Modeling},
  author={Liu, Xiaoyu and Liang, Di and Shan, Hongyu and Liu, Peiyang and Liu, Yonghao and Wu, Muling and Li, Yuntao and Wu, Xianjie and Miao, Li and Shen, Jiangrong and others},
  booktitle={Proceedings of the 2025 Conference on Empirical Methods in Natural Language Processing: Industry Track},
  pages={672--685},
  year={2025}
}

@article{liu2025stole,
  title={Who Stole Your Data? A Method for Detecting Unauthorized RAG Theft},
  author={Liu, Peiyang and Cui, Ziqiang and Liang, Di and Ye, Wei},
  journal={arXiv preprint arXiv:2510.07728},
  year={2025}
}

@article{dai2025hope,
  title={HoPE: Hyperbolic Rotary Positional Encoding for Stable Long-Range Dependency Modeling in Large Language Models},
  author={Dai, Chang and Shan, Hongyu and Song, Mingyang and Liang, Di},
  journal={arXiv preprint arXiv:2509.05218},
  year={2025}
}

\end{document}